\documentclass[letterpaper, 10 pt, conference, nofonttune]{ieeeconf} 

\IEEEoverridecommandlockouts                              

\usepackage{amsthm}

\usepackage{subfigure}
\usepackage{xcolor}
\usepackage{times}
\usepackage{epsfig}
\usepackage{amsmath}
\usepackage{amssymb}
\usepackage{mathrsfs}
\usepackage[ruled,vlined,linesnumbered]{algorithm2e}
\usepackage{wrapfig}
\usepackage{multirow}
\usepackage{microtype}
\usepackage{booktabs}
\usepackage{adjustbox}
\usepackage{cite}
\usepackage{graphicx}

\usepackage{hyperref}
\usepackage{lineno}

\newcommand{\GGG}{\mathcal{G}}

\newcommand{\HHH}{\mathcal{H}}

\newcommand{\LLL}{\mathcal{L}}

\newcommand{\EE}{\mathbf{E}}

\newcommand{\vv}{\mathbf{v}}
\newcommand{\kk}{\mathbf{k}}

\newcommand{\vel}{\mathbf{q}}
\newcommand{\uu}{\mathbf{u}}

\newcommand{\WW}{\mathbf{W}}

\newcommand{\zz}{\mathbf{z}}
\newcommand{\ff}{\mathbf{f}}

\newcommand{\xx}{\mathbf{x}}

\makeatother

\title{\LARGE \bf
Collaborative Planning with Concurrent Synchronization for \\Operationally Constrained UAV-UGV Teams
}

\author{Zihao Deng$^{1}$, Qianhuang Li$^{1}$, Peng Gao$^{2}$,
Maggie Wigness$^{3}$, John Rogers$^{3}$, Donghyun Kim$^{1}$, and Hao Zhang$^{1}$
\thanks{*This work was partially supported by NSF CAREER Award IIS-2308492, DEVCOM ARL TBAM CRA W911NF2520024, and DARPA Young Faculty Award (YFA) D21AP10114-00.}
\thanks{$^{1}$Zihao Deng, Qianhuang Li, Donghyun Kim, and Hao Zhang are with the University of Massachusetts Amherst, Amherst, MA, 01002, USA.
Email: \{zihaodeng, qianhuangli, donghyunkim, hao.zhang\}@umass.edu.}%
\thanks{$^{2}$Peng Gao is with North Carolina State University, Raleigh, NC, 27695, USA. {Email: pgao5@ncsu.edu}.}%
\thanks{$^{3}$Maggie Wigness and John Rogers are with the U.S. Army DEVCOM Army Research Laboratory (ARL), Adelphi, MD, 20783, USA. Email: \{maggie.b. wigness, john.g.rogers59\}.civ@army.mil.}%
}

\begin{document}

\maketitle
\thispagestyle{empty}
\pagestyle{empty}

\begin{abstract}
Collaborative planning under operational constraints is an essential capability for heterogeneous robot teams tackling complex large-scale real-world tasks. Unmanned Aerial Vehicles (UAVs) offer rapid environmental coverage, but flight time is often limited by energy constraints, whereas Unmanned Ground Vehicles (UGVs) have greater energy capacity to support long-duration missions, but movement is constrained by traversable terrain. Individually, neither can complete tasks such as environmental monitoring.
Effective UAV-UGV collaboration therefore requires energy-constrained multi-UAV task planning, traversability-constrained multi-UGV path planning, and crucially, synchronized concurrent co-planning to ensure timely in-mission recharging.
To enable these capabilities, we propose \textit{Collaborative Planning with Concurrent Synchronization} (CoPCS), a learning-based approach that integrates a heterogeneous graph transformer for operationally constrained task encoding with a transformer decoder for joint, synchronized co-planning that enables UAVs and UGVs to act concurrently in a coordinated manner. CoPCS is trained end-to-end under a unified imitation learning paradigm.
We conducted extensive experiments to evaluate CoPCS in both robotic simulations and physical robot teams. Experimental results demonstrate that our method provides the novel multi-robot capability of synchronized concurrent co-planning and substantially improves team performance.

More details of this work are available on the project website:
\href{https://hcrlab.gitlab.io/project/CoPCS/}{https://hcrlab.gitlab.io/project/CoPCS}.
\end{abstract}


\section{Introduction}

Multi-robot systems are a promising paradigm in robotics because of their advantages such as redundancy \cite{gao2023collaborative}, parallelism \cite{pinciroli2012argos}, and scalability \cite{portugal2013multi}.
These characteristics make multi-robot systems indispensable for addressing complex, large-scale tasks that are usually infeasible for a single robot.
Collaborative multi-robot planning is a fundamental capability that allows robots to collaborate on their actions, share resources, and complete interdependent tasks under operational constraints.
Effective collaborative planning in multi-robot systems is necessary not only for achieving operational efficiency but also for ensuring the reliability of robot teams when deployed in a variety of real-world applications, including disaster response \cite{erdelj2016uav, khan2022emerging, erdelj2017wireless} and environmental monitoring \cite{asadzadeh2022uav, liu2022uav, fascista2022toward, deng2025coordinated, dengsubteaming}.

\begin{figure}[!t]
\vspace{6pt}
\centering
\includegraphics[width=0.48\textwidth]{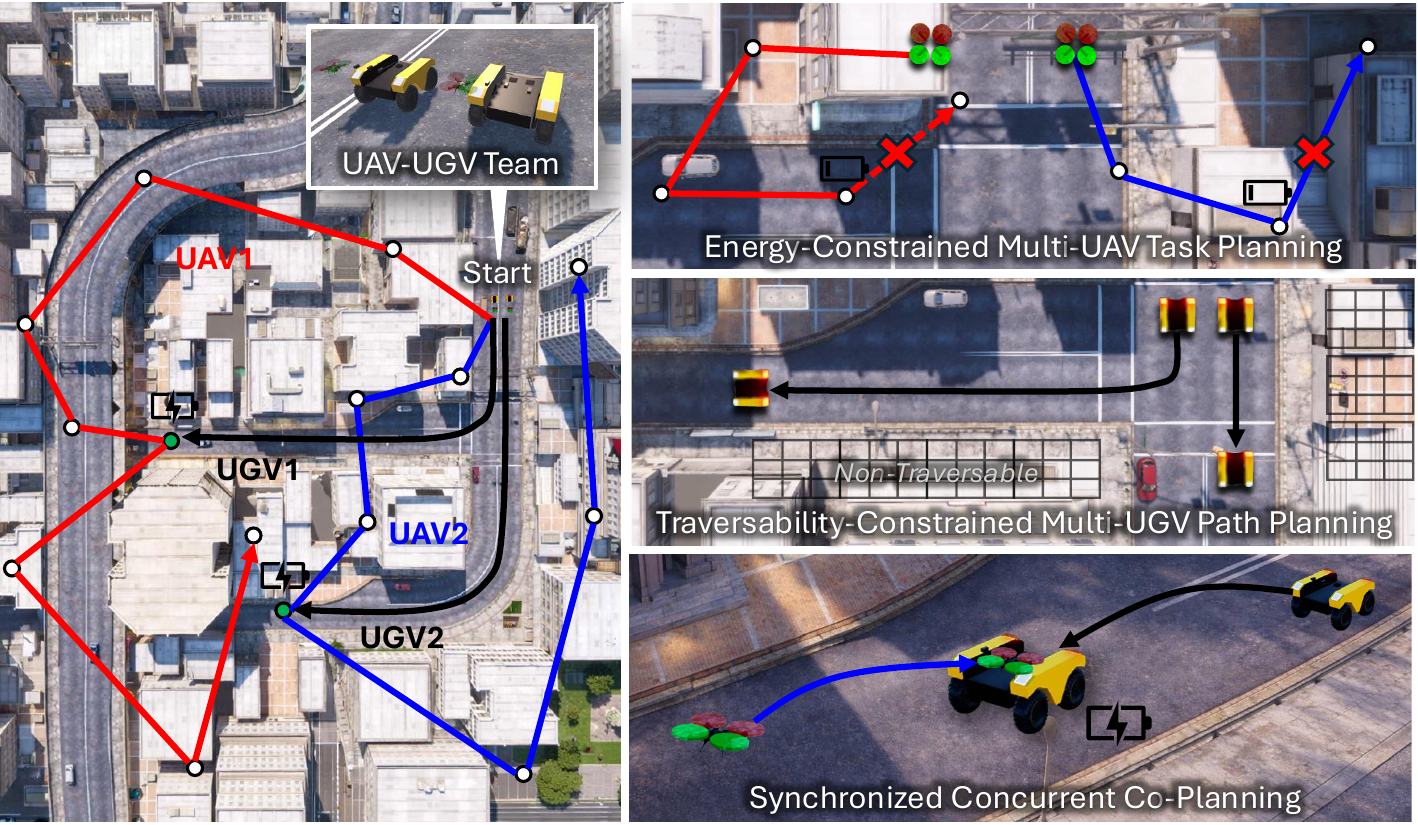}
\caption{A motivating scenario of a UAV-UGV team performing collaborative planning under operational constraints to visit all task points for environmental monitoring. UAVs are flight-time limited due to energy constraints, while UGVs are constrained by terrain traversability, preventing either platform from completing the mission alone. To overcome this, UGVs act as mobile recharging stations, carrying energy supplies and synchronizing with UAVs to sustain their operations. Task points are shown as white circles and recharging points as green circles, while colored lines denote UAV trajectories and black lines denote UGV paths.
}
\label{fig:motivation}
\end{figure}

Multi-robot systems often consist of heterogeneous robots with diverse capabilities and operational constraints.
For example, Unmanned Aerial Vehicles (UAVs) excel at providing rapid environmental coverage and reaching locations that are inaccessible or difficult for ground robots.
However, their endurance is severely constrained by limited energy capacity, which restricts both the scale and duration of their missions. In contrast, Unmanned Ground Vehicles (UGVs) are often slower and subject to terrain traversability constraints; however, they can carry larger energy reserves and function as mobile recharging stations for UAVs.
Figure \ref{fig:motivation} illustrates a scenario in which a UAV-UGV team collaborates to visit all task points for environment monitoring. 
Neither UAVs nor UGVs alone can accomplish the mission due to energy and traversability limitations; 
however, UGVs can collaborate with UAVs and sustain their operations through in-mission recharging. 
In such cases, UAVs must determine which task locations to visit and the visit sequence given their energy constraints (i.e., \textit{energy-constrained multi-UAV task planning}). 
UGVs must plan paths along traversable road networks that support UAV operations (i.e., \textit{traversability-constrained multi-UGV path planning}). 
Most importantly, these plans must be jointly optimized and time-synchronized to guarantee timely in-mission recharging (i.e., \textit{synchronized concurrent co-planning}).

Due to the importance of UAV-UGV collaborative planning, various methods have been explored. 
Traditional approaches, including Mixed Integer Programming \cite{murray2015flying}, Dynamic Programming \cite{liang2024algorithm}, and metaheuristics \cite{arnold2019knowledge}, 
can usually identify optimal solutions. 
However, these methods often demand substantial domain knowledge and can become computationally expensive for large-scale problems, limiting their applicability for real-time execution.
Learning-based methods address these limitations and have shown promising results. 
For example, 
graph neural networks are used to capture the relationships between robots and tasks \cite{luo2022graph},
imitation learning allows systems to mimic expert demonstrations \cite{javarone2016evolutionary}, 
and
reinforcement learning improves decision-making through environmental feedback \cite{mondal2025coordinate}.
However, most existing learning-based methods fail to adequately address operational constraints,
and often fail to achieve synchronized concurrent co-planning between UAVs and UGVs, 
which can result in longer mission times or even mission failure.

To address these challenges, we propose the method of \textit{Collaborative Planning with Concurrent Synchronization} (CoPCS) for operationally constrained UAV-UGV teams,
which enables a UAV-UGV team to work collaboratively: 
energy-constrained UAVs plan their task-point visits, 
traversability-constrained UGVs plan supportive routes, 
and the full team achieves concurrent synchronization to ensure timely in-mission recharging.
Specifically, we design a heterogeneous graph representation to encode entities, including tasks, path points, and heterogeneous robots. Each node represents an entity with its attributes, and each edge captures the spatial relationships among the entities. Given this graph, CoPCS uses a heterogeneous graph transformer to generate constraint-aware embeddings that facilitate synchronized collaboration. 
Using these embeddings, CoPCS employs a transformer decoder to generate a joint action sequence for both UAVs and UGVs through co-planning under operational constraints, covering task visits, path movements, and recharging decisions. The unified framework is trained end-to-end using imitation learning, and UAV and UGV actions are executed simultaneously, resulting in synchronized concurrent co-planning.

Our key contribution is the proposal of the CoPCS approach for collaborative planning with concurrent synchronization for operationally constrained UAV-UGV teams. 
Specific novelties of this work are as follows:

\begin{itemize}

\item We enable a new multi-robot capability for constrained heterogeneous co-planning, 
which simultaneously addresses energy-constrained multi-UAV task planning, traversability-constrained multi-UGV path planning, and most importantly, synchronized concurrent co-planning for the entire team.

\item We introduce CoPCS as one of the first unified learning-based solutions that integrates heterogeneous graphs for operationally constrained task encoding with transformer decoders for sequential and synchronized action generation, which is trained end-to-end under a unified imitation learning paradigm.

\end{itemize}

\section{Related Work}\label{sec:related}
\subsection{Classic Methods for UAV-UGV Collaborative Planning}
A variety of traditional approaches have been implemented to address the collaborative multi-robot planning problem, and can be broadly categorized into optimization-based and heuristic-based approaches. Optimization methods formulate UAV-UGV collaborative planning as NP-hard problems such as the Vehicle Routing Problem with Drones (VRP-D) or the Traveling Salesman Problem with Drones (TSP-D). 
Mixed Integer Programming (MIP) employs binary and continuous variables to encode task assignments and vehicle routes \cite{murray2015flying, manyam2019cooperative, di2017last, climaco2021branch}, while Dynamic Programming (DP) recursively decomposes the problem into subproblems (e.g., shortest paths over subsets of tasks) and combines their solutions using the principle of optimality \cite{liang2024algorithm, kool2022deep, lera2022dynamic}. Although these methods provide exact solutions, their computational cost grows rapidly with problem size, which limits scalability.

To address this, heuristic and metaheuristic methods have been widely adopted. Guided Local Search (GLS) \cite{arnold2019knowledge}, Tabu Search (TS) \cite{gmira2021tabu}, and Simulated Annealing (SA) \cite{yaugmur2021multi} iteratively improve candidate solutions by exploring neighborhoods while using different strategies to escape local optima. In UAV-UGV collaborative planning, they are often implemented in multi-level frameworks, where UGV routes are planned first and UAV task assignments optimized afterward \cite{maini2015cooperation, ropero2019terra, mondal2023optimizing}. While this decomposition can yield optimal solutions within each level, it cannot guarantee global optimality across the mission. Compared to optimization methods, heuristics and metaheuristics reduce computational costs and scale better to large problem instances, but they typically cannot ensure global optimality and often remain too slow for real-time execution.

\subsection{Learning for UAV-UGV Collaborative Planning}
Learning-based approaches have recently shown strong potential to address the computational limitations of optimization methods. 
Graph Neural Networks (GNNs) capture relational structures between tasks and agents through message passing  \cite{prates2019learning, luo2022graph, boffa2022neural}. 
Imitation Learning (IL) complements this by mimicking expert demonstrations to generalize solutions \cite{javarone2016evolutionary}.
Reinforcement Learning (RL) learns decision-making policies via interactions with the environments, 
An attention-based network has been applied to solve the TSP and extended to multi-TSP \cite{zhang2023coordinated, fuertes2023solving}.
However, these methods neglect practical constraints in real-world applications, such as UAV energy. 
To address this, several studies introduce energy-constrained UAV planning with fixed recharging stations \cite{betalo2025dynamic, fan2022deep, brotee2024optimizing}, but they mainly focus on homogeneous UAV teams without considering collaboration with UGVs.

To improve efficiency, recent efforts explore UAV-UGV co-planning, where UGVs act as mobile recharging stations to eliminate UAV detours to fixed stations \cite{mondal2025risk, mondal2024attention}. 
Yet, these studies are limited to single UAV-UGV settings. 
More recently, co-planning multiple energy-constrained UAVs and UGVs has introduced the capability to coordinate larger teams \cite{mondal2025coordinate}.
However, these approaches often rely on greedy strategies to assign UAVs to UGVs and lack concurrent synchronization across the team. As a result, UAVs and UGVs cannot execute their plans simultaneously, leading to unsuccessful missions or longer completion times, since UAVs may be forced to wait for UGVs to arrive before recharging.

\section{Approach} \label{sec:approach}
\subsection{Problem Definition}


\begin{figure*}[h]
\includegraphics[width=0.975\textwidth]{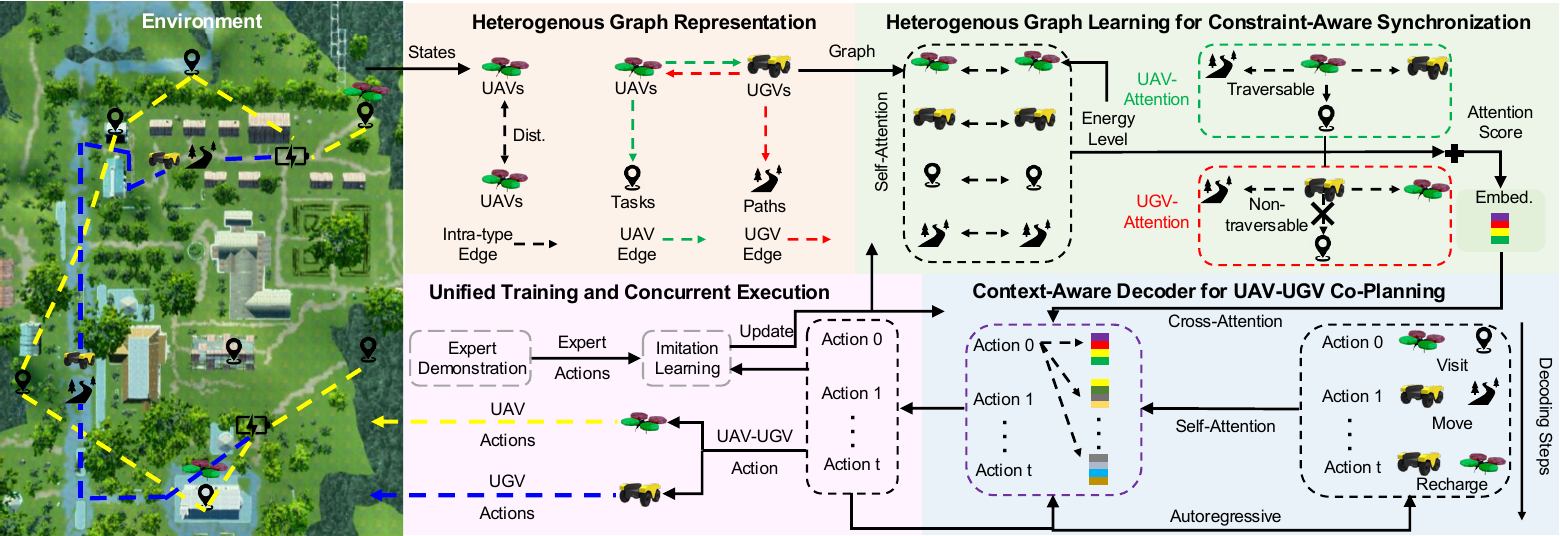}
\centering
\caption{Overview of CoPCS (\textit{Collaborative Planning with Concurrent Synchronization}), a unified learning-based method that integrates a heterogeneous graph transformer for synchronized collaboration among operationally constrained UAV-UGV teams and a transformer decoder for joint action co-planning. The CoPCS approach is trained end-to-end, and its co-planning policy is executed concurrently to coordinate all the robots, which enables concurrent synchronization. 
}\label{fig:architecture}
\end{figure*}

We discuss our CoPCS approach that enables the new multi-robot capabilities of synchronized concurrent co-planning for heterogeneous robot teams.  An overview of CoPCS is illustrated in Figure \ref{fig:architecture}. 
We denote the set of task points as $\mathcal{M} = \{m_0, m_1, \ldots, m_{\mid \mathcal{M}\mid-1}\}$,
which must be visited by a team of energy-constrained UAVs, defined as  $\mathcal{U}^a = \{u^a_0, u^a_1, \ldots, u^a_{\mid \mathcal{U}^a\mid-1}\}$.
In addition, we denote the team of UGVs as $\mathcal{U}^g = \{u^g_0, u^g_1, \ldots, u^g_{\mid \mathcal{U}^g\mid-1}\}\}$,
which must plan their paths over traversable terrains
to reach a set of points $\mathcal{P} = \{p_0, p_1, \ldots, p_{\mid \mathcal{P}\mid-1}\}$ in order to recharge the UAVs.

Formally, we represent the environmental context as a graph $\mathcal{G} = \{\mathcal{T},\mathcal{E}^{intra}, \mathcal{E}^{uav}, \mathcal{E}^{ugv}\}$.
The node set $\mathcal{T}$ is defined as $\mathcal{T} = \mathcal{T}^{m} \cup \mathcal{T}^{p} \cup \mathcal{T}^{a} \cup \mathcal{T}^{g}$. 
The task set $\mathcal{T}^{m} = \{\mathbf{t}_i^{m}\}$ encodes the locations of all tasks, where $\mathbf{t}_i^{m} = [\mathbf{d}_i^m, \delta_i]$, with $\mathbf{d}_i^m$ denoting the two-dimensional coordinates of the task location along the x and y directions, and $\delta_i\in\{0,1\}$ indicating whether the task location has been visited or not.  
The path set $\mathcal{T}^{p} = \{\mathbf{t}_i^{p}\}$ contains the central locations of all paths, where  $\mathbf{t}_i^{p}$ denotes the position of the central location of the $i$-th path. 
The UAV set $\mathcal{T}^{a} = \{\mathbf{t}_i^{a}\}$ contains the attributes of UAVs, where $\mathbf{t}_i^{a} = [\mathbf{d}_i^a, f_i]$. 
The UAV position is denoted as $\mathbf{d}_i^a$, and $f_i$ indicates the remaining energy. 
The UGV set $\mathcal{T}^{g} = \{\mathbf{t}_i^{g}\}$ contains the attributes of UGVs, where $\mathbf{t}_i^{g} =[\mathbf{d}_i^g, \mathbf{d}^a]$, with $\mathbf{d}_i^g$ denoting the UGV position, and $\mathbf{d}^a$ indicating the position of the UAVs. 

We define three types of edge connectivity based on relative distance. 
The intra-type edge set is denoted as $\mathcal{E}^{intra} =  \{e^{intra}_{i,j}\}$, which connects nodes of the same type. 
The UAV edge set is defined as $\mathcal{E}^{uav} =  \{e^{uav}_{i,j}\}$, which connects UAV nodes with their related nodes, including UGVs and tasks. 
Similarly, the UGV edge set is denoted as $\mathcal{E}^{ugv} =  \{e^{ugv}_{i,j}\}$, which connects UGV nodes with their related nodes, including UAVs and paths. 
In all cases, an edge $e_{i,j}=1$ if the $i$-th and $j$-th nodes are within a predefined radius; otherwise $e_{i,j}=0$.

Given the environmental context graph $\GGG$, we aim to collaboratively plan the actions of UAVs and UGVs. 
Formally, the UAV action is defined as $\mathbf{A}^{uav}=\{a_{i,j}\}^{|\mathcal{U}^a|\times |\mathcal{M}|}$, where $a_{i,j}=1$ indicates that the $i$-th UAV visits the $j$-th task. The UGV action is defined as $\mathbf{A}^{ugv}=\{a_{i,j}\}^{|\mathcal{U}^g|\times |\mathcal{P}|}$, where $a_{i,j}=1$ indicates that the $i$-th UGV selects the $j$-th path to traverse. In addition, the recharging action is defined as $\mathbf{A}^{recharge}=\{a_{i,j}\}^{|\mathcal{U}^a|\times |\mathcal{U}^g|}$, where $a_{i,j}=1$ indicates that the $i$-th UAV is recharging on the $j$-th UGV.
Our goal is to address the following UAV-UGV collaborative planning problems under operational constraints:
\begin{itemize}
    \item \textbf{Synchronized concurrent co-planning:} The joint planning of UAV and UGV actions to achieve synchronized movement, which ensures simultaneous arrival at charging locations, as well as concurrent operation to enable UAVs and UGVs to operate in parallel for continuous and efficient task execution.
    
    \item \textbf{Energy-constrained multi-UAV task planning:} The coordination of multiple UAVs with limited energy so that all task points are visited exactly once while minimizing the overall mission completion time.

    \item \textbf{Traversability-constrained multi-UGV path planning:} The coordination of multiple UGVs to plan feasible and efficient routes over a road network, subject to terrain traversability constraints.
\end{itemize}

\subsection{Heterogeneous Graph Learning for Constraint-Aware Synchronization}
Given the environmental context graph $\GGG$,
we develop a heterogeneous graph transformer network \cite{hu2020heterogeneous} $\psi$ to compute the node embedding as $\mathcal{H} = \{ \mathbf{h}_i \} = \psi(\mathcal{G})$, 
where $\mathbf{h}_i$ denotes the embedding of the $i$-th node. 
These embeddings capture the relationships among UAVs, UGVs, tasks, and paths, which enable constraint-aware synchronization that further facilitates collaborative planning.

The network $\psi$ first projects the node attributes $\mathbf{t}_i$ into a feature vector $\mathbf{z}_i$ of the $i$-th node using linear layers by $\mathbf{z}_i = \mathbf{W}^{z} \mathbf{t}_i$, 
where $\WW^{z}$ denotes the type-specific learnable weight matrix of the linear layer. 
Then, we compute the query, key, and value embeddings of each node by:
\begin{equation}
    \vel_i^{l} = \WW_q^{l} \zz_i^{l}, \quad
    \kk_i^{l} = \WW_k^{l} \zz_i^{l}, \quad
    \vv_i^{l} = \WW_v^{l} \zz_i^{l}
\end{equation}
where $\vel_i^{l}$, $\kk_i^{l}$, and $\vv_i^{l}$ denote query, key, and value at the $l$-th layer. $\WW_q^{l}$, $\WW_k^{l}$,  and $\WW_v^{l}$ denote their weight matrices. 

To encode relationships among nodes of the same type (e.g., within the UAV team), we apply a self-attention mechanism that aggregates features across nodes of that type. The self-attention score from node $j$ to connected node $i$ in $\mathcal{E}^{intra}$ at layer $l$ is computed as follows: 
\begin{equation}
\alpha_{i,j}^{l} = \text{SoftMax}\left( \frac{ (\mathbf{q}_i^{l})^\top \mathbf{k}_j^{l} }{ \sqrt{d} } \right)
\end{equation}
where SoftMax($\cdot$) is the softmax function and $d$ is the dimension of the $\vel_i^l$, $\kk_j^l$.

To facilitate synchronization among heterogeneous robots, we employ a cross-attention mechanism that aggregates features across selected node types based on their edge connectivity.
For UAVs, message passing is performed along all UAV edges $\mathcal{E}^{uav}$, and the cross-attention score from the connected $k$-th node to the $i$-th UAV node at layer $l$ is defined as:
\begin{equation}
\beta_{i,k}^{l, a} = \text{SoftMax}\left( \frac{ (\mathbf{q}_i^{l})^\top \mathbf{k}_k^{l} }{ \sqrt{d} } \right)
\end{equation}
For UGVs, we explicitly encode traversability constraints by restricting message passing to UGV edges $\mathcal{E}^{ugv}$ that exclude connections to task nodes, since tasks may lie in non-traversable regions. 
Formally, the cross-attention score from the connected $n$-th node to the $i$-th UGV node in at layer $l$ is defined as:
\begin{equation}
\beta_{i,n}^{l, g} = \text{SoftMax}\left( \frac{ (\mathbf{q}_i^{l})^\top \mathbf{k}_n^{l} }{ \sqrt{d} } \right)
\end{equation}

Finally, given the scores from both self-attention and cross-attention mechanisms, we update the embedding $\mathbf{h}_i$ of each node by aggregating both scores, which is defined as follows: 

{\footnotesize
\begin{equation}
\mathbf{h}_i^{l+1}=\mathrm{LN}\!\Big(
\mathbf{h}_i^{l}
+\!\!\sum_{e_{i,j}^{intra}=1}\!\alpha_{i,j}^{l}\mathbf{v}_{j}^{l}
+\!\!\sum_{e_{i,k}^{uav}=1}\!\beta_{i,k}^{l,a}\mathbf{v}_{k}^{l}
+\!\!\sum_{e_{i,n}^{ugv}=1}\!\beta_{i,n}^{l,g}\mathbf{v}_{n}^{l}
\Big)
\end{equation}
}where $\text{LN}(\cdot)$ denotes the layer normalization operation. 
The updated embedding $\mathbf{h}_i^{l+1}$ for the $i$-th node encodes both intra-type features (second term) and inter-type features (third and fourth term). This explicitly enables synchronization because robots are able to exchange information with one another, while remaining constraint-aware through the inclusion of UAV energy and UGV traversability in their respective embeddings.

\begin{figure*}[t]
\includegraphics[width=1\textwidth]{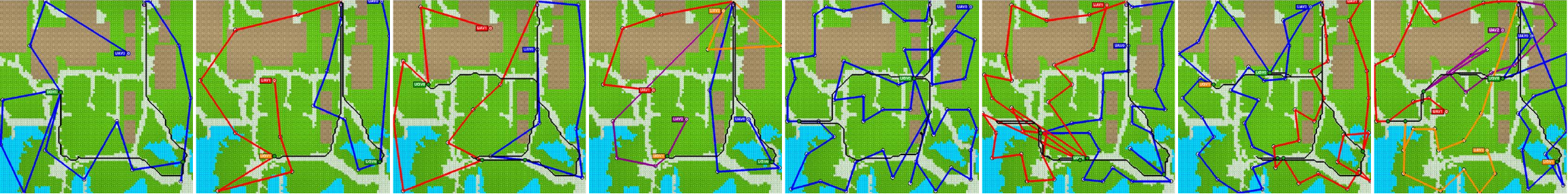}
\centering
\vspace{-12pt}
\caption{Qualitative results of CoPCS in a map-based simulator.
The first four figures illustrate a mission with 15 tasks and 5 paths, while the last four figures show a mission with 45 tasks and 10 paths. The experiments evaluate different team configurations, ranging from 1 UAV and 1 UGV to 4 UAVs and 2 UGVs. Task points are depicted as white circles, recharging points as green circles, UAV trajectories as colored lines, and UGV paths as black lines.}\label{fig:grid}
\end{figure*}

\begin{table*}[tb]
\centering
\caption{Quantitative results under different mission requirements across operationally constrained UAV-UGV team configurations.}
\label{tb:combined}
\begin{adjustbox}{max width=\textwidth}
\begin{tabular}{llcccccccccccc}
\toprule
\textbf{Mission} & \textbf{Method} & \multicolumn{3}{c}{\textbf{1UAV--1UGV}} & \multicolumn{3}{c}{\textbf{2UAV--1UGV}} & \multicolumn{3}{c}{\textbf{2UAV--2UGV}} & \multicolumn{3}{c}{\textbf{4UAV--2UGV}} \\
\cmidrule(lr){3-5} \cmidrule(lr){6-8} \cmidrule(lr){9-11} \cmidrule(lr){12-14}
& & Makespan & UAV energy & UGV energy & Makespan & UAV energy & UGV energy & Makespan & UAV energy & UGV energy & Makespan & UAV energy & UGV energy \\
\midrule
\multirow{5}{*}{T15-P5} 
& GLS \cite{arnold2019knowledge} & 8662.71 & 1050.57 & 13262.04 & 7108.87 & 1115.01 & 13472.43 & 6319.81 & 1068.90 & 21972.07 & 5522.37 & 1322.88 & 24697.03 \\
& TS \cite{gmira2021tabu}  & 9086.49 & 1010.59 & 13906.42 & 6390.69 & 1059.31 & 12872.67 & 8119.87 & 1328.26 & 19737.92 & 6770.71 & 1389.42 & 28746.32 \\
& SA \cite{yaugmur2021multi} & 9078.41 & 1032.61 & 14639.16 & 6860.22 & 1177.55 & 14502.63 & 7125.12 & 1320.02 & 17222.68 & 5238.74 & 1321.54 & 18715.35 \\
& MLP \cite{park2023learn} & 6936.44 & 934.34 & 10927.01 & 4043.61 & 1070.81 & 8512.61 & 4983.65 & 1167.34 & 12353.34 & 3182.47 & 1327.13 & 11051.53 \\
& CoPCS & \textbf{6617.75} & \textbf{875.92} & \textbf{10016.21} & \textbf{3477.51} & \textbf{934.75} & \textbf{8251.61} & \textbf{3465.41} & \textbf{965.06} & \textbf{11511.28} & \textbf{2614.08} & \textbf{1180.29} & \textbf{8569.27} \\
\midrule
\multirow{5}{*}{T45-P10} 
& GLS \cite{arnold2019knowledge} & 15073.27 & 1836.16 & 22369.79 & 12628.09 & 2385.66 & 23615.11 & 8884.95 & 2136.77 & 22014.14 & 8143.29 & 2728.93 & 29281.86 \\
& TS \cite{gmira2021tabu}  & 16160.50 & 1978.89 & 24510.74 & 10543.42 & 2387.07 & 21277.91 & 11110.28 & 2251.52 & 22086.83 & 9617.13 & 2901.18 & 32365.98 \\
& SA \cite{yaugmur2021multi} & 18075.23 & 2116.41 & 25632.18 & 16719.52 & 2692.45 & 37677.11 & 10507.05 & 2444.45 & 21901.16 & 9536.52 & 2626.84 & 33265.91 \\
& MLP \cite{park2023learn} & 14326.98 & 2295.39 & 12793.25 & 8213.82 & 1886.78 & 11854.86 & 8351.97 & 1808.43 & 16052.93 & 6394.29 & 2224.99 & 17242.63 \\
& CoPCS & \textbf{11900.02} & \textbf{1744.71} & \textbf{9707.79} & \textbf{6659.41} & \textbf{1719.32} & \textbf{11249.41} & \textbf{6379.37} & \textbf{1769.12} & \textbf{11742.68} & \textbf{5973.65} & \textbf{2098.53} & \textbf{15077.62} \\
\bottomrule
\end{tabular}
\end{adjustbox}
\end{table*}

\subsection{Context-Aware Decoder for UAV-UGV Co-Planning}
Given the embedding set $\mathcal{H}$ generated by the heterogeneous graph transformer network $\psi$, we develop a transformer-based decoder \cite{vaswani2017attention} $\phi$ to generate joint action sequences for both UAVs and UGVs in an auto-regressive manner, 
thereby allowing co-planning. 
The decoder predicts one action at each decoding step $t$, it generates action $a_t$ conditioned on the action history $\{a_1, \dots, a_{t-1}\}$ and the graph embedding $\mathcal{H}$, where $a_t \in \mathbf{A}^{uav}$ if it is a UAV visiting action,  $a_t \in \mathbf{A}^{ugv}$ if it is a UGV moving action, and $a_t \in \mathbf{A}^{recharge}$ if it is a recharge action.
This autoregressive process begins with an empty sequence and continues generating actions until mission completion, enabling coherent co-planning through modeling of temporal dependencies and environmental context.

Specifically, at each decoding step $t$, the decoder network $\phi$ first projects each action $a_i$ in the current action sequence and its positional index $i$ into a feature vector $\mathbf{x}_i$, where $i \leq t-1$. 
This process is defined as $\mathbf{x}_i=\EE^a[a_i]+\EE^i[i]$, where $\EE^a$ and $\EE^i$ are learnable embedding matrices for actions and positional indices, respectively. 
To generate coherent actions for the robot team, we apply a causal self-attention mechanism to model temporal dependencies, which ensures that the generation of action $a_t$ only attends to the previous actions $\{a_1, \dots, a_{t-1}\}$. 
The network $\phi$ projects the feature vector $\mathbf{x}_i$ into query, key, and value embeddings, which are defined as follows:
\begin{equation}
    \vel_i^{l,o} = \WW_q^{l,o} \xx_i^{l}, \; 
    \kk_i^{l,o} = \WW_k^{l,o} \xx_i^{l}, \; 
    \vv_i^{l,o} = \WW_v^{l,o} \xx_i^{l}
\end{equation}
where $\vel_i^{l,o}$, $\kk_i^{l,o}$, and $\vv_i^{l,o}$ denote query, key, and value at the $l$-th layer for each action $a_i$, $\WW_q^{l,o}$, $\WW_k^{l,o}$, and $\WW_v^{l,o}$ denote their learnable weight matrices. 
Given these embeddings, the self-attention score is computed as follows:
\begin{equation}
    \gamma_{i,j}^{l} = \text{SoftMax}\left( \frac{ (\mathbf{q}_i^{l,o})^\top \mathbf{k}_j^{l,o} }{ \sqrt{d} } \right)
\end{equation}
where $\gamma_{i,j}^{l}$ denotes the attention score from action $a_j$ to $a_i$. 

To enable context-aware action generation for collaborative planning, we condition each action on the graph embeddings from $\mathcal{H}$, which includes the contextual information over the environment. 
Specifically, the network $\phi$ projects the action feature $\xx_i$ into query and graph embedding $\mathbf{h}_k$ into key and value, which is defined as follows:
\begin{equation}
    \vel_i^{l,c} = \WW_q^{l,c} \xx_i^{l}, \; 
    \kk_k^{l,c} = \WW_k^{l,c} \mathbf{h}_k, \; 
    \vv_k^{l,c} = \WW_v^{l,c} \mathbf{h}_k
\end{equation}
where $\vel_i^{l,c}$, $\kk_k^{l,c}$, and $\vv_k^{l,c}$ denote query, key, and value at the $l$-th layer, $\WW_q^{l,c}$, $\WW_k^{l,c}$, and $\WW_v^{l,c}$ denote their learnable weight matrices.
Given these embeddings, the cross-attention score is computed as follows:
\begin{equation}
    \delta_{i,k}^{l} = \text{SoftMax}\left( \frac{ (\mathbf{q}_i^{l,c})^\top \mathbf{k}_k^{l,c} }{ \sqrt{d} } \right)
\end{equation}

We further define the hidden representation $\mathbf{f}_i^l$ for each action $a_i$, which serves as its contextual embedding at the $l$-th decoder layer. 
Starting with $\mathbf{f}_i^0 = \mathbf{x}_i$, the hidden representation is iteratively updated by aggregating both self-attention and cross-attention, defined as follows:
\begin{equation}\small
\mathbf{f}_i^{l+1} = \text{LN} \left( \mathbf{f}_i^{l} +
\sum_{j \leq i} \gamma_{i,j}^{l} \mathbf{v}_j^{l,o} +
\sum_{k \in \mathcal{H}} \delta_{i,k}^{l} \mathbf{v}_k^{l,c} \right)
\end{equation}
where the second term captures temporal dependencies among actions through self-attention, and the third incorporates contextual information from the graph embeddings through cross-attention. 
Finally, the network $\phi$ projects the final hidden representation $\mathbf{f}_{t-1}^{L}$ from the last layer into action logits by $\uu_t=\WW^u\ff_{t-1}^L$, where $\WW^u$ denotes the learnable matrix. 
The action probability distribution conditioned on the graph embedding is then computed as $P(a_t\mid a_1, a_2, \dots, a_{t-1}, \HHH)=\text{SoftMax}(\uu_t)$ and the action is then selected greedily as:
\begin{equation}
    a_t = \arg\max_{a_t} P(a_t\mid a_1, a_2, \dots, a_{t-1}, \mathcal{H})
\end{equation}
The selected action $a_t$ is appended to the action history, forming $\{a_1, a_2, \dots, a_t\}$, and the decoder uses this new sequence to generate the next action $a_{t+1}$. This autoregressive process continues until all tasks are completed.

\subsection{Unified Training and Concurrent Execution}
Since our approach requires learning complex collaborative policies for heterogeneous robots, we need substantial training data covering diverse mission scenarios. 
Manual labeling of optimal action sequences for task visiting, path movement, and recharging across all scenarios is impractical.
Therefore, we adopt imitation learning using demonstrations from a Mixed Integer Programming (MIP) solver that generates optimal solutions for each training scenario. 
The MIP solver formulates the UAV-UGV collaboration as a constrained optimization problem that enables multi-UAV task planning and multi-UGV path planning. 
The MIP solver generates optimal action sequences as ground truth actions $\{a_1^*, a_2^*, \dots, a_{t}^*\}$. 
While the MIP solver provides optimal solutions, it has exponential time complexity and cannot execute in real time for planning.

Given the demonstration as ground truth, we train the network $\psi$ and $\phi$ in a unified way using an autoregressive training algorithm with teacher forcing. 
Specifically, the algorithm receives a 2D occupancy map of the environment including the task points $\mathcal{M}$ and path points $\mathcal{P}$, as well as the UAVs $\mathcal{U}^a$ and UGVs $\mathcal{U}^g$ along with their attributes $\mathbf{t}_i$. 
The network $\psi$ processes the $\mathbf{t}_i$ into embeddings $\mathcal{H}$.
The network $\phi$ generates the action sequence conditioned on the embedding $\mathcal{H}$. 
At each step $t$, the decoder receives the ground truth previous action $\{a_1^*, a_2^*, \dots, a_{t-1}^*\}$ to predict the probability distribution over all possible actions, and then generates the action $a_t$.  
The training objective is to minimize the cross-entropy loss between predicted and ground truth actions, which is defined as follows:
\begin{equation}
    \LLL=-\sum_{t=1}^T\log P(a_t=a_t^* \mid a_1^*, a_2^*, \dots, a_{t-1}^*, \HHH)
\end{equation}
where the loss is minimized using Adam as the optimization solver \cite{kingma2017adammethodstochasticoptimization}. 
The gradients computed from this loss are backpropagated through the transformer decoder $\phi$ to the heterogeneous graph transformer $\psi$, which enables unified end-to-end training of all learnable parameters for context-aware embedding generation and sequential action prediction. 
The training time complexity is dominated by $O(BL_dL^2)$ per iteration, where $B$ is the batch size, $L_d$ is the number of decoder layers, and $L$ is the action sequence length. 

During execution, our approach assumes a 2D occupancy map of the environment, including tasks, paths, UAVs, and UGVs, consistent with the training setup. 
The heterogeneous graph transformer network generates embeddings for each entity, providing constraint-aware representations that enable synchronized collaboration within the UAV-UGV team. 
Based on these embeddings, the context-aware transformer decoder produces a joint action sequence that ensures coherent co-planning of UAVs and UGVs. 
With these preconditions, 
this joint sequence is then partitioned into UAV-specific and UGV-specific action sequences, which are executed concurrently, allowing both vehicle types to operate in parallel and achieve synchronized concurrent co-planning.
The execution time complexity is dominated by $O(L_dL^2)$ for each scenario. 
\section{Experiments}

\begin{figure*}[t]
    \centering
    \includegraphics[width=1\textwidth]{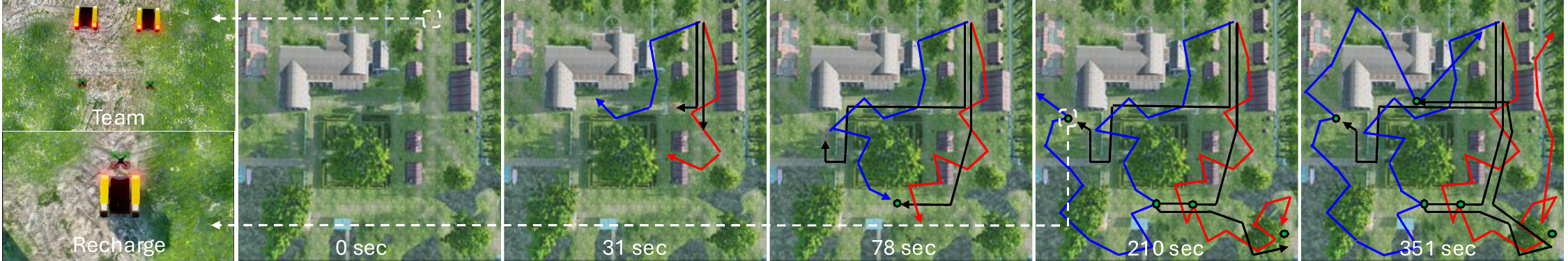}
    \vspace{-12pt}
    \caption{Qualitative results of co-planning with concurrent synchronization for a team of two Warthog UGVs and two quadrotor UAVs in a high-fidelity Unity-based 3D simulator integrated with ROS1. The mission includes 45 tasks and 10 traversable paths. The first column of the two subfigures illustrates the team starting configuration and a synchronized recharging event during execution, while the remaining subfigures depict trajectories at different timestamps. }
    \label{fig:flood}
\end{figure*}

\begin{figure*}
    \centering
    \includegraphics[width=1\textwidth]{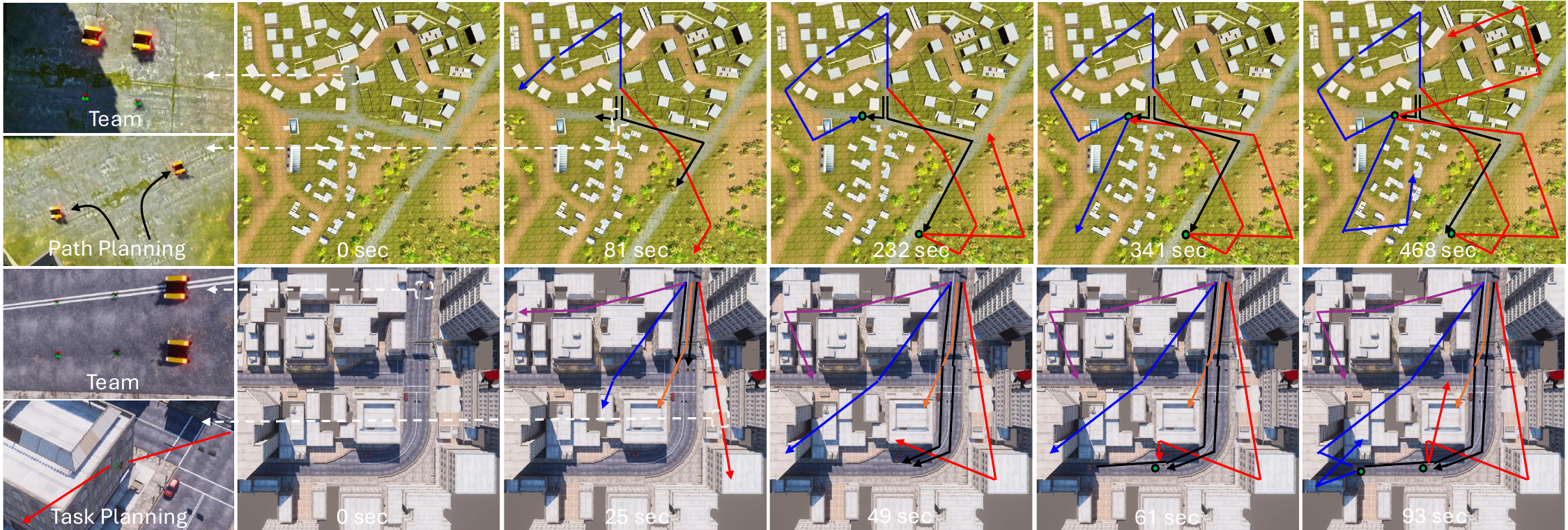}
    \vspace{-12pt}
    \caption{Qualitative results illustrating the generalizability of CoPCS in unseen environments. The large-scale suburban and urban maps are not used during training and also contain variations in starting points and road network layouts. In the top environment, a 2-UAV and 2-UGV team is required to visit 15 tasks, while in the bottom environment, a 4-UAV and 2-UGV team is required to visit 15 tasks.}
    \label{fig:gene}
\end{figure*}

\subsection{Experimental Setups}
We comprehensively evaluate CoPCS in three setups: a map-based 2D simulator, a Unity 3D multi-robot simulator with ROS1, and physical robot teams with ROS2. Evaluations span multiple team configurations (from 1 UAV-1 UGV to 4 UAV-2 UGV) and varying mission complexities, defined by the number of tasks and available UGV paths (15 tasks/5 paths and 45 tasks/10 paths).
Simulations
experiments are conducted within a $10\text{km} \times 10\text{km}$ operational area. 
UAVs operate at a constant speed of $q_a = 10$ m/s, while UGVs traverse road networks at $q_g = 4.5$ m/s. 
Each UAV has an energy capacity of $ 287.7$ kJ, while each UGV has an energy capacity of $36,810$ kJ. The energy consumption of UAVs and UGVs follows the robot profiles defined in \cite{hurwitz2021mobile}, expressed as $0.0461(q_a)^3-0.5834(q_a)^2-1.8761q_a+229.6$ and $464.8q_g+356.3$, respectively. 
In each scenario, the team begins from a common depot. Energy-constrained UAVs visit designated task points, while traversability-constrained UGVs provide recharging support along road networks. 

We generate synthetic datasets to train and evaluate our approach under varying team configurations and mission requirements. The differences between datasets lie in the number of robots, the number of tasks to be completed, and the number of paths available. Each dataset contains 1,000 problem instances, with task positions randomly generated. Within each dataset, 800 instances are used for training, while the remaining 200 instances are reserved for testing and evaluation.

We compare our CoPCS approach with three metaheuristic methods which focus on fixed UGV path and a baseline method:
(1) Guided Local Search (\textbf{GLS}) \cite{arnold2019knowledge} that improves UAV task routes by penalizing frequently used task assignments to escape local optima and explore alternative routing strategies, 
(2) Tabu Search (\textbf{TS}) \cite{gmira2021tabu} that explores UAV task routes while maintaining memory of recent moves to prevent cycling back to previous solutions, 
(3) Simulated Annealing (\textbf{SA}) \cite{yaugmur2021multi} that initially accepts suboptimal UAV task routes with high probability, then gradually becomes more selective as the search converges, 
and 
(4) Multilayer Perceptron (\textbf{MLP}) \cite{park2023learn} that uses the same transformer network $\phi$ for sequential planning, but uses a multilayer perceptron to generate the embeddings $\mathcal{H}$. 
To quantitatively evaluate and compare our approach with baseline methods, we employ four metrics for collaborative planning. 
(1) \textbf{Makespan} \cite{10207838} is defined as the total mission completion time (i.e., the time required to visit all task points) measured in seconds (s), 
(2) \textbf{UAV energy} is the cumulative energy consumption of all UAVs measured in kilojoules (kJ), 
(3) \textbf{UGV energy} is the cumulative energy consumption of all UGVs measured in kilojoules (kJ). 

\begin{figure*}
    \centering
    \includegraphics[width=1\textwidth]{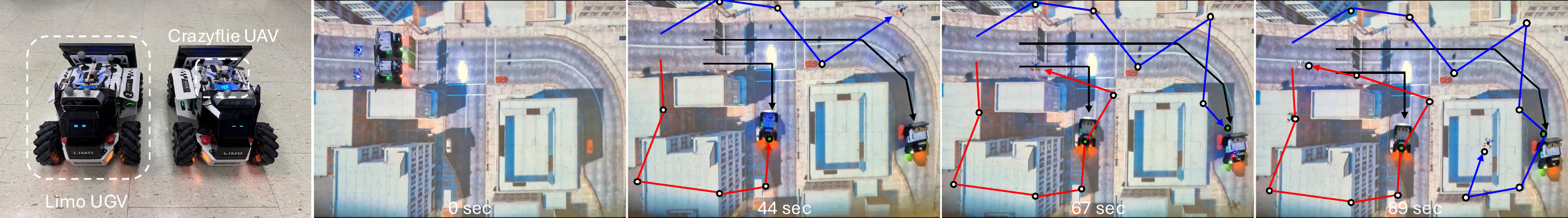}
    \vspace{-12pt}
    \caption{Qualitative results of CoPCS in real-world mixed-reality experiments using two Crazyflie UAVs and two Limo UGVs controlled through ROS2.  }
    \label{fig:real}
\end{figure*}

\subsection{Results in Map-Based Robotics Simulations}
We first evaluate CoPCS in map-based simulations to explore its capability for collaborative planning under different mission requirements and UAV-UGV team configurations.

The qualitative results in the map-based 2D simulator are shown in Figure \ref{fig:grid}. The results demonstrate that CoPCS enables operationally constrained UAV-UGV teams to collaborate effectively and accomplish the environmental monitoring task. 
We noticed that CoPCS introduces synchronized concurrent co-planning, allowing UAVs and UGVs to operate in parallel for continuous execution while ensuring synchronization at recharging points, as illustrated by the recharging behavior at the 210-second timestamp. It also supports multi-UAV task planning by determining which tasks each UAV should visit and in what order, reducing makespan under energy constraints. In addition, CoPCS coordinates multi-UGV path planning over road networks while respecting traversability constraints.

The quantitative results are presented in Table~\ref{tb:combined}.
For the mission with 15 tasks, the three metaheuristic baselines exhibit the weakest performance, particularly in makespan and UGV energy consumption, as they optimize only UAV task planning while relying on fixed UGV paths, thereby neglecting synchronized collaboration.
The MLP baseline performs better; however, its simple architecture fails to capture complex environment structure, resulting in longer makespan and higher energy consumption.
In contrast, CoPCS jointly plans UGV paths and UAV trajectories under synchronized collaboration, achieving the lowest energy consumption and the shortest makespan.
When the mission scales to 45 tasks, the performance gap between CoPCS and the baselines widens substantially.
The metaheuristic methods perform the worst, and the MLP baseline improves slightly but remains limited.
CoPCS consistently outperforms all baselines across all team configurations, achieving shorter makespan and lower energy consumption in this larger and more complex mission by enabling synchronized concurrent planning.

\subsection{Results Using High-Fidelity 3D Simulations with ROS1}
We  validate CoPCS in a high-fidelity Unity-based 3D simulator integrated with ROS1 for robot control. Qualitative results are shown in Figure \ref{fig:flood}. The simulator replicates outdoor field environments for environmental monitoring tasks. 
The experiments involve two wheeled Warthog UGVs constrained by terrain features such as water and buildings, and two quadrotor UAVs limited by energy capacity. 
The team configuration is illustrated in the first subfigure. CoPCS enables UAVs and UGVs to operate in parallel: the UAV team allocates tasks and determines visit sequences, while the UGVs coordinate to plan feasible routes. More importantly, the UAV-UGV team achieves concurrent synchronization, allowing timely recharging and ensuring successful mission completion. One such recharging event is shown in the second subfigure. These results demonstrate the effectiveness of CoPCS in handling collaborative planning for UAV-UGV teams.


We further evaluate the generalizability of CoPCS in unseen environments, as illustrated in Figure \ref{fig:gene}. 
CoPCS is trained on datasets generated from the rural environment shown in Figure~\ref{fig:flood}, and its generalization is assessed on two additional maps.
The first map represents a suburban setting where a 2UAV-2UGV team is required to visit 15 tasks, while the second map depicts an urban environment where a 4UAV-2UGV team must complete 15 tasks. 
These maps present distinct scenarios with different starting points and road network layouts. 
In both cases, CoPCS enables the team to visit all task points successfully. 
Specifically, in the suburban map, the two UGVs coordinate to plan supportive paths in different directions to recharge UAVs, while in the urban map, the UAVs effectively plan their task visits to maximize efficiency. Overall, the results highlight the capability of our approach to generalize to diverse mission scenarios.

\subsection{Case Studies on Physical UAV-UGV Teams with ROS2}

We validate CoPCS in real-world case studies using a UAV-UGV team composed of two Crazyflie UAVs and two Limo UGVs.
The team operates in a mixed-reality setup, where a projector renders a simulated outdoor city scene onto the physical environment and a motion capture system tracks robot poses as well as task and path locations.
The team is centrally controlled via ROS2 from a master workstation equipped with an NVIDIA RTX 4090 GPU and 64 GB RAM.
Qualitative results are shown in Figure~\ref{fig:real}. Tasks are distributed across the map, including non-traversable regions for the Limo robots, while the third and fourth subfigures illustrate team recharge events.
In this setup, the CoPCS policy achieves an inference rate of 9 Hz for 15 tasks and 3 Hz for 45 tasks, which is sufficient for our real-time planning requirements.
Overall, the results demonstrate the effectiveness and applicability of our approach in enabling physical robot teams to achieve collaborative planning with concurrent synchronization.

\section{Conclusion}\label{sec:conclusion}
In this paper, we propose CoPCS, a framework for collaborative planning with concurrent synchronization in operationally constrained UAV-UGV teams. 
CoPCS integrates heterogeneous graph learning for constraint-aware synchronization with a context-aware decoder for co-planning, enabling UAVs and UGVs to act concurrently within a unified framework trained end-to-end under an imitation learning paradigm.
Results from comprehensive experiments demonstrate that CoPCS introduces the new multi-robot capability for constrained UAV-UGV teams of synchronized concurrent co-planning, and significantly outperforms existing methods. 
Future work will explore decentralized extensions via distributed graph representations and local message passing to improve scalability and robustness in larger multi-robot teams.

\bibliographystyle{IEEEtran}
\bibliography{ref}
\end{document}